\documentclass[letterpaper, 10 pt, conference]{ieeeconf}  

\IEEEoverridecommandlockouts                              

\usepackage{graphicx}

\usepackage{xcolor}
\usepackage{array}
\usepackage{float}
\usepackage{booktabs}
\usepackage{adjustbox}
\usepackage[noadjust]{cite}
\usepackage{multirow}
\usepackage{marginnote}

\usepackage[a-1b]{pdfx}

\title{\LARGE \bf
A Fabric-based Pneumatic Actuator for the Infant Elbow: Design and Comparative Kinematic Analysis}

\author{Ipsita Sahin,$^{1}$  Mehrnoosh Ayazi,$^{2}$ Caio Mucchiani,$^{2}$ Jared Dube,$^{1}$ Konstantinos Karydis,$^{2}$ and Elena Kokkoni$^{1}$
\thanks{$^{1}$Dept. of Bioengineering; $^{2}$Dept. of Electrical and Computer Engineering, University of California, Riverside. 
Email:{\tt\footnotesize\{isahi001, mayaz004, caiocesr, jdube004, karydis, elenak\}@ucr.edu}. 
We gratefully acknowledge the support of NSF award \# CMMI-2133084. 
Any opinions, findings, and conclusions or recommendations expressed in this material are those of the authors and do not necessarily reflect the views of the National Science Foundation.
}}

\begin{document}
\maketitle
\thispagestyle{empty}
\pagestyle{empty}

\begin{abstract}

This paper focuses on the design and systematic evaluation of fabric-based, bellow-type soft pneumatic actuators to assist with flexion and extension of the elbow, intended for use in infant wearable devices. 
Initially, the performance of a range of actuator variants was explored via simulation. 
The actuator variants were parameterized based on the shape, number, and size of the cells present. 
Subsequently, viable actuator variants identified from the simulations were fabricated and underwent further testing on a physical model based on an infant's body anthropometrics.
The performance of these variants was evaluated based on kinematic analyses using metrics
including movement smoothness, path length, and elbow joint angle. Internal pressure of the actuators was also attained. Taken together, results reported herein provide valuable insights about the suitability of several actuator designs to serve as components for pediatric wearable assistive devices. 
\end{abstract}

\section{Introduction}
Certain conditions (e.g., arthrogryposis, cerebral palsy, etc.) may impede the ability to perform upper extremity (UE) motor activities early in life~\cite{oishi2017treatment, bunata2014cerebral}.
The use of assistive devices can help pediatric populations with functional impairments engage in such activities, leading to gains in motor function~\cite{Henderson2008}.
The availability of assistive devices specifically for pediatric populations under the age of two years is limited~\cite{arnold2020exploring}, with only a few UE devices being under development~\cite{Kokkoni2020_asme} and/or being tested with infants~\cite{babik2016feasibility, lobo2016playskin}.

There has been an increasing interest in the use of soft robotics toward the development of UE assistive devices~\cite{shahid2018moving, majidi2021review, thalman2020review, pan2022soft}. 
Compared to their rigid counterparts, soft robotics can lead to more lightweight and low-profile yet functional designs which helps increase their safety~\cite{Xiong2020}. 
While earlier related work focused on the design of soft actuators to support motion about smaller UE joints, such as those of the fingers and wrist joints~\cite{polygerinos2015soft, shahid2018moving}, currently there exist systems that support motion of larger and more complex joints, such as the shoulder and elbow joints~\cite{ang2020design, oguntosin2015development}. 
However, these systems have been developed primarily for adults. Very recent efforts have focused on infants~\cite{Kokkoni2020_asme, sahin2022bidirectional, mucchiani2022closed, gollob2023length, Paez2021, yamamoto2015wearable}, and out of these efforts, only one design has specifically focused on the elbow joint~\cite{Kokkoni2020_asme}. 
Providing motion about the elbow joint early in life is important as it is involved in reaching, bringing objects to the mouth for exploration, and eating, among other activities~\cite{berthier2006development}. 

Assistance in elbow flexion and extension (albeit for devices focused on adults) has been primarily achieved by soft bellow-type actuators~\cite{ang2020design, thalman2018novel, huang2023low,oguntosin2015development, nassour2021soft}. 
These actuators consist of a system of air pouches (aka cells) which are connected together in series (in most cases) to form an array. 
Different fabrication methods and materials have been used to make these type of actuators. Most notable examples include casted silicone actuators, and actuators made of some form of heat-sealable fabric such as flexible thermoplastic polyurethane (TPU). 
This work considers the latter type of fabric-based actuators. 

There are key advantages to the use of fabric-based bellow-type actuators in the context of this work. 
They are compliant, lightweight and of low profile, thus posing a low risk of causing injuries in case of malfunction, while at the same time they have a high power-to-weight ratio. 
Compared with other engineered devices~\cite{ Xiong2020}, these actuators are more scalable owing to their lower cost and easier fabrication~\cite{majidi2021review, polygerinos2017soft}. 
Additionally, bellow-type designs can achieve greater directional deflection compared to flat-membrane soft actuators of the same size~\cite{yang1997micro,wu2008kinematically}. 
Last yet importantly, their scalability feature makes devices comprising fabric-based bellow type actuators resonate well with the inherent need for a device focused on infant populations to adapt to fast growth. Indeed, as the infant develops and gains strength, the number of cells of the actuators can also be adjusted to provide the appropriate level of assistance, ensuring customized and progressive training~\cite{gollob2023length, natividad2018reconfigurable}. 
%

In our previous work, we have developed an actuated UE wearable device for infants using pneumatically-controlled silicone-based actuators to provide assistance at both elbow and shoulder joints~\cite{Kokkoni2020_asme}. We also explored the use of fabric-based actuators to support shoulder abduction/adduction~\cite{sahin2022bidirectional, mucchiani2022closed} and flexion/extension~\cite{mucchiani2023robust}. 
In this work, we investigate the utility of fabric-based soft pneumatic actuators as an upgraded alternative to the silicone-based ones previously used to assist elbow flexion/extension. 
\section{Actuator Design, Fabrication, and Preliminary Feasibility Testing}




The paper introduces a new family of multi-cell bellow-type actuators made of TPU fabric. 
First-principles-based numerical simulation was conducted to estimate stress-displacement profiles of different actuator designs varying in cell shape and number of cells.
Designs deemed viable from simulations were fabricated and experimentally tested for their performance on a physical model. 

\subsection{Design Considerations}

The choice of fabric material is critical for ensuring the desired compliance and functionality of the actuators. 
Typically, elastomeric fabrics with high tensile strength, compliant material characteristics, and elasticity are used (such as TPU films)~\cite{natividad2018reconfigurable}. 
%
Flexible and lightweight fabric (Oxford 200D heat-sealable coated fabric of $0.20$\;mm thickness) was used herein.
The fabric can be heat-sealed via non-specialized equipment like a household iron, a straightener, or a flat iron. 

\begin{figure}[!t]
\vspace{6pt}
     \centering
     \includegraphics[width=0.75\columnwidth]{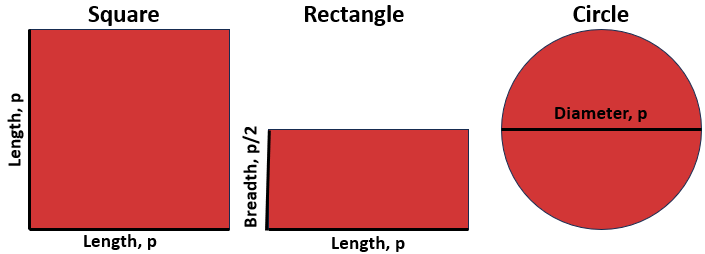}
     \vspace{-10pt}
         \caption{Actuator cell shapes considered herein, along with the corresponding key dimension, cell length $p\in\{1, 2, 3, 4\}$\;cm. The square cell has the largest inner cavity while the rectangle cell has the smallest. }
     \label{designs1}
     \vspace{-18pt}
\end{figure}

Our actuator design was based on three key characteristics: i) cell shape (i.e. square, rectangle, and circle), ii) cell length (Fig.~\ref{designs1}), and iii) the number of cells~\cite{ma2023modeling, yang2018new, sheng2020multi, natividad2018reconfigurable}.
The shape, length, and number of cells determine the inner space to be filled, which in turn impacts the inflation/deflation duration and thus influences the motion and characteristics of the bellow type actuator~\cite{ma2023modeling}. 
Actuator dimensions were first narrowed down based on anthropometric data of the target infant population. 
The upper arm and forearm lengths for infants of 6-12 months of age vary on average between $13.8-16.4\;$cm and $8.57-10.86\;$cm, respectively, while the circumference varies between $13.10-14.74\;$cm and $12.74-14.55\;$cm, respectively~\cite{Fryar2021, edmond2020normal}. 
Given that an infant's arm diameter is between $4.05-4.69\;$cm, we ruled out actuator designs with maximum cell dimension over $4$\;cm. 

\subsection{Preliminary Feasibility}

\begin{figure}[!t]
\vspace{6pt}
     \centering
     \includegraphics[width=0.48\textwidth]{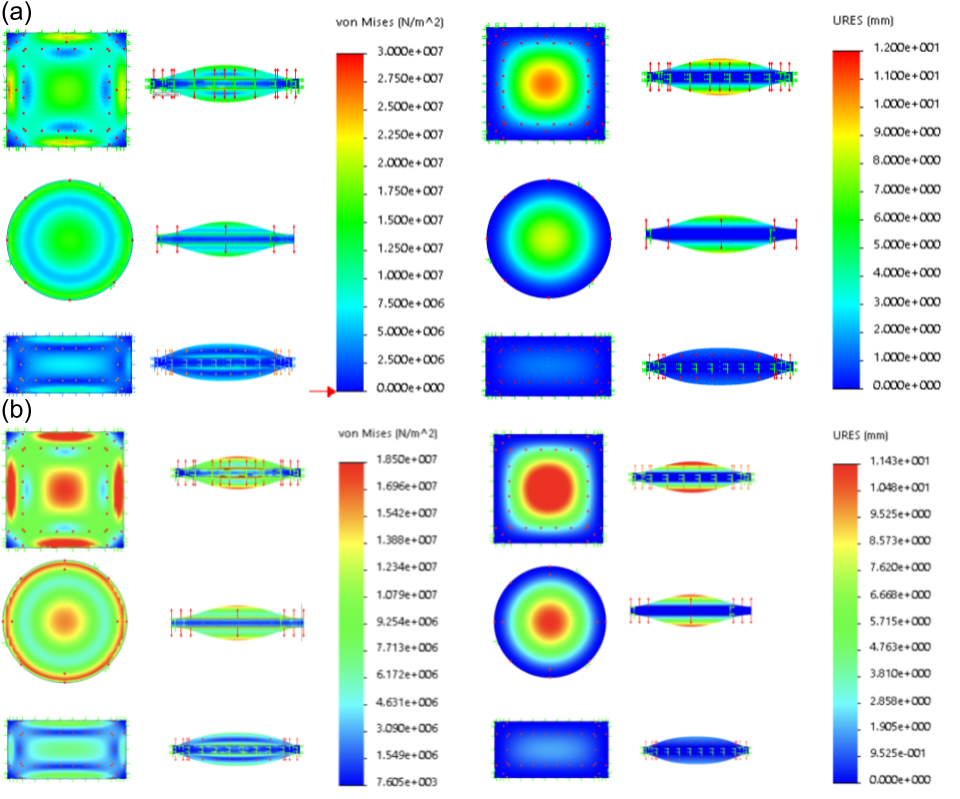}
     \vspace{-16pt}
         \caption{Sample stress (panels in left) and displacement (panels in right) profiles (both top and side views are shown) for one cell actuator variant of (a) 3cm and (b) 4cm (not in scale) in length actuators. Stress decreases and displacement increases with larger cell lengths.}
     \label{fig: simulation}
     \vspace{-12pt}
\end{figure}

The performance of one cell of varying shape and length (Fig.~\ref{designs1}) was first assessed through simulation (SolidWorks). 
Simulations were used to estimate the stress and displacement for each cell variant (Fig.~\ref{fig: simulation}). 
Results show that the maximum displacement can be attained at the cell center. 
This suggests that the central area of each cell should be connected to the following one to maximize elongation during (multi-cell) actuator inflation. 
Also, single-cell elongation increases with larger cell length while stress decreases. 
This finding supports the use of larger cell-length actuators.
We further used single-cell displacement values to estimate the maximum elongation of different multi-cell actuators consisting of the aforementioned actuator cell variants (rectangle, square and circle cell of length $p\in\{1,2,3,4\}$\;cm). 
We considered cases with number of cells, $n\in\{1,6,8,10,12,14\}$. 
This led to 72 distinctive actuator variants. 
In all cases the cell thickness was $0.50$\;mm when at rest. 
We assumed a static case where each actuator's free end can elongate along a single direction without any obstruction. 


\begin{figure}[!t]
\vspace{6pt}
     \centering
     \includegraphics[width=1\columnwidth]{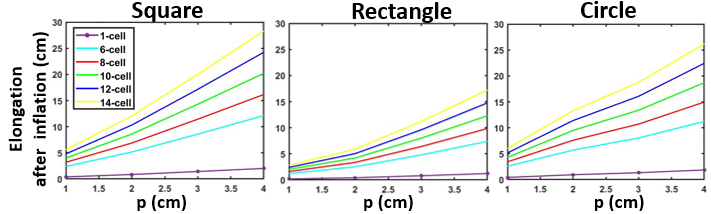}
     \vspace{-18pt}
        \caption{Elongation increases as cell length $p$ increases. The rectangle showed comparatively smaller elongation compared to the respective square-shaped cells since the inner cavity of the former is about half that of the square cell. No noteworthy differences in terms of elongation ranges were observed between square- and circle-shaped cells.}
     \label{designs2}
     \vspace{-15pt}
\end{figure}

\begin{figure}[!t]
\vspace{6pt}
     \centering
     \includegraphics[width=0.85\columnwidth]{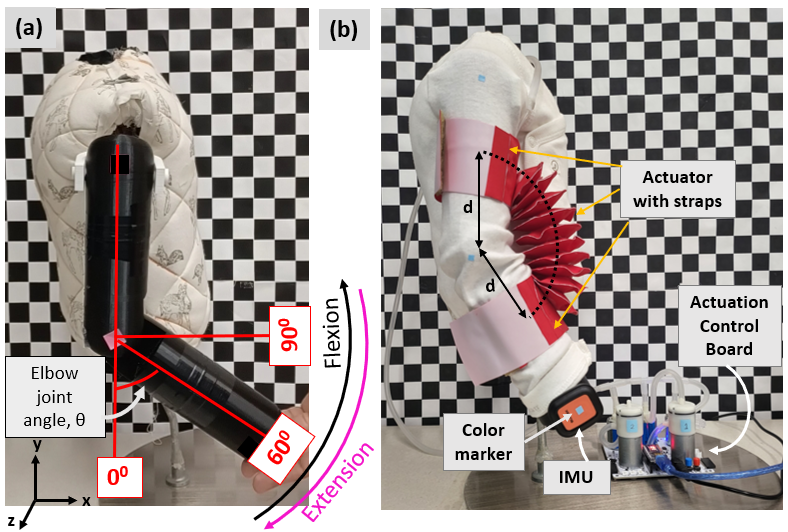}
     \vspace{-10pt}
         \caption{Experimental setup for the physical experiments.}
     \label{setup}
     \vspace{-10pt}
\end{figure}

Figure~\ref{designs2} depicts those estimated values. 
Adding more cells not only makes the actuator bulkier (e.g., 14-cell) but it also increases the inflation/deflation time to values prohibitive for our application.
\footnote{~To provide more context, the time taken by an infant to complete a reaching action does not exceed two seconds~\cite{zhou2021infant}.}  
On the other hand, the maximum feasible elongation for the lower number of cells (e.g., 6-cell) and/or smaller cell lengths (e.g., $1$\;cm) may tightly hold the arms creating discomfort for the infant when worn. 
A design consideration of our prototype wearable device is for the elbow actuator to be placed on the top of the elbow and symmetrically about the elbow joint (Fig.~\ref{setup}). 
Hence, the actuator should be able to cover the arc length of a circle with a radius equal to the distance, $d$, between the elbow joint and the attachment of the actuator to the upper/forearm. 
This distance should not exceed $5$\;cm from the elbow joint as the center of mass for the forearm of infants lies approximately $45\%$ from the proximal joint~\cite{schneider1992mass}. 
Thus, to initiate the elbow joint at $\ge90$ degrees elbow joint 
the actuator needs to elongate for at least $8$\;cm. 
Considering this approximation, the 6-cell actuator for all of the selected dimensions may not be able to provide the minimum required performance. 

\begin{figure}[!t]
\vspace{6pt}
     \centering
     \includegraphics[width=0.485\textwidth]{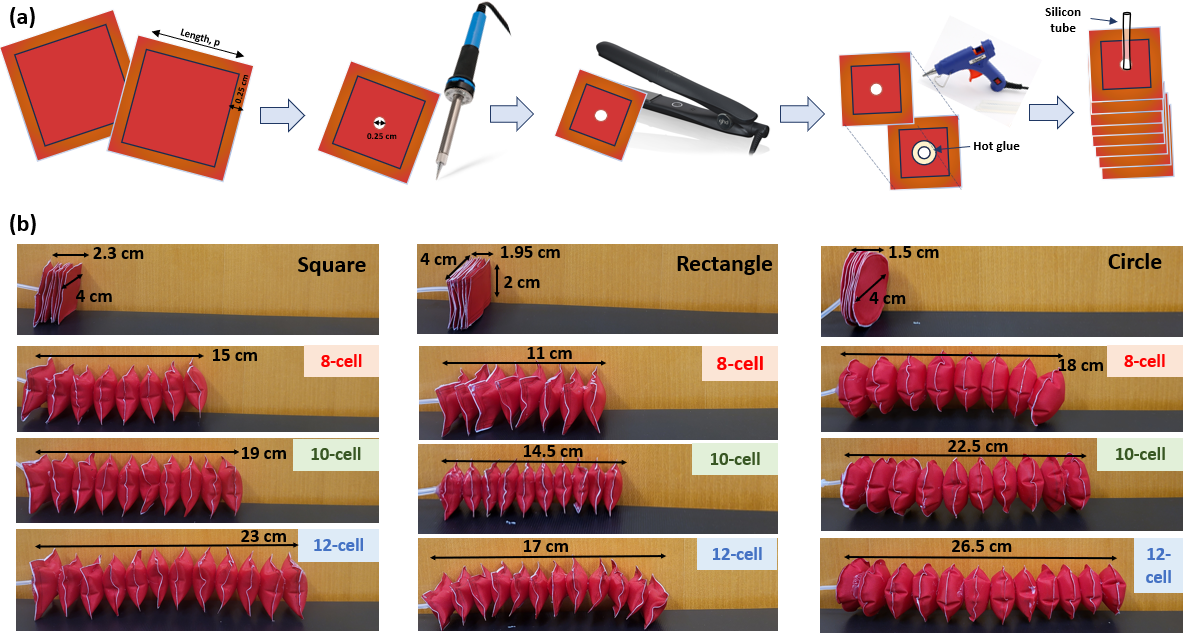}
     \vspace{-20pt}
         \caption{(a) Fabrication process of the proposed actuator. The process contains five consecutive steps: cut out the two sides, make the air channel using a soldering iron, heat-seal edges using a flat iron, attach cells with hot glue, and finally attach the silicon tube. 
         (b) $4$\;cm cell length actuators of varying shape and number of cells in deflated and fully inflated conditions. }
     \label{actuator}
     \vspace{-15pt}
\end{figure}

Based on these findings and design considerations, 
we elected to retain as viable all 8-, 10- and 12-cell actuator variants with cell length of $2$, $3$, and $4$\;cm. 
All these actuator variants were fabricated and tested in static physical experiments. 
All the actuator variants were fabricated following the steps outlined in Fig.~\ref{actuator}a and checked individually for any air leakage to ensure proper functionality. 
Fabrication time for each variant was about 1 to 1.5 hours. 
The corners of the proposed geometric shapes require a meticulous heat-sealing process during fabrication since these are more prone to leakage and popping. 
The smaller the cell size, the harder to effectively heat seal the edges or periphery without specialized equipment. 
The circle-shaped actuators were thus considerably more challenging to fabricate since heat-sealing circles is not as trivial as the edges of a square or rectangle. 

We tested all these actuator variants in static experiments (Fig.~\ref{actuator}b). 
Experimental and simulated results are consistent. 
The actuators can elongate for several times their length when at rest ($5.4\times - 11\times$), without any bending as pressure is being applied, and when the free end is not obstructed. As discussed next, the actuators can also bend and support rotational motion (i.e. flexion/extension) when placed around the elbow, thus making them multi-functional. (In this work, forearm supination/pronation was not considered.)

\begin{table}[!t]
    \vspace{6pt}
    \caption{Estimated and Experimental Elongation}
    \label{compare}
    \vspace{-12pt}
    \begin{center}
        \begin{adjustbox}{width=0.9\columnwidth,center}
            \begin{tabular}{c c c c c c c c}
                \toprule
                \multirow{3}{*}{Size} & \multirow{3}{*}{Shape} & \multicolumn{6}{c}{Elongation (cm)}\\
                & & \multicolumn{2}{c}{8-cell} & \multicolumn{2}{c}{10-cell} & \multicolumn{2}{c}{12-cell} \\
                & & Est. & Expt. & Est. & Expt.& Est. & Expt. \\
                \midrule
                \multirow{3}{*}{3cm} & Square & 11.44 & 12.00 & 14.30 & 14.70& 17.16 & 18.00\\
                & Rectangle & 6.40 & 7.00 & 8.00 & 9.00 & 9.60& 10.65\\
                & Circle & 10.72 & 11.00 & 13.40 & 13.50 & 16.08 & 17.00\\
                \midrule
                \multirow{3}{*}{4cm} & Square & 16.16 & 15.00 & 20.20 & 19.00& 24.24 & 23.00\\
                & Rectangle & 9.84 & 11.00 & 12.30 & 14.50 & 14.76& 17.00\\
                & Circle & 14.96 & 18.00 & 18.70 & 22.50 & 22.44 & 26.50\\
                \bottomrule
            \end{tabular}
        \end{adjustbox}
    \end{center}
        \vspace{-25pt}
\end{table}

The experimentally-observed elongation values also generally match the predicted values (Table~\ref{compare}), although the circular actuators were observed to exhibit greater elongation compared to their square or rectangle-shaped counterparts. 
Actuator weight varies between $3.5-28\;$g, making them suitable for wearable robotics applications. 
Physical experiments demonstrated that actuators with cell lengths of $2$\; cm (especially the circle and rectangle shapes) are quite challenging to manufacture with significant variability among different cells and with frequent air leakage. 
Physical static experiments enabled us to deduce that $2$\;cm cell length actuators are possible, but to improve consistency their manufacturing process has to adapt. 
The latter is part of ongoing work and out of the scope of this paper, hence all $2$\;cm cell length actuators were excluded.

The final 18 distinct actuator variants deemed as the most viable choices for elbow flexion/extension with lower chances of failure include all actuators combining cell numbers of $\{8,10,12\}$, cell lengths of $\{3,4\}$\;cm, and cell shapes of $\{$square, rectangle, circle$\}$. 
These were further tested experimentally as discussed in the next section. 


\section{Testing and Evaluation}

\subsection{Experimental Setup}
For the physical experiments, the downselected actuator variants were placed on a physical model scaled approximately to the 50\textsuperscript{th} percentile of a 12-month-old infant's upper body~\cite{Fryar2021, edmond2020normal} (Fig.~\ref{setup}).
Two straps of $16$\;cm adjustable in length were fixed to both actuator ends to secure it onto the arm.
Inflation and deflation of the actuators were controlled using an off-body pneumatic control board (Programmable-Air) detailed in~\cite{sahin2022bidirectional}.  
The 3D-printed hollow arm was filled with sandbags for the correct weight ($\sim0.38$\;kg) and the upper arm and forearm lengths measured $15$\;cm and $11$\;cm, respectively, matching anthropomorphic measurements for the specified age range.
The physical model's elbow joint had a passive range of motion (ROM) of $105$\;degrees.

Strategic placement of the actuators was chosen to enable elbow flexion and extension while preserving rotational movement along other axes. 
The actuator was symmetrically placed on the top of the elbow joint, i.e. the top half placed on and the top end attached to the upper arm, and the bottom half placed on and the bottom end attached to the forearm (Fig.~\ref{setup}b).
The actuator's deflation contributes to elbow flexion, while its inflation assists with elbow extension.

The actuators on the infant model underwent a series of 10 trials involving inflation and deflation cycles of five seconds. 
In the experiments, we operated the actuation control board at 100\% PWM (Pulse Width Modulation) as it offers the quickest inflation/deflation compared to other PWM levels. 
Video recordings from the experiments were analyzed to determine the 2D positions of the end-effector, elbow, and shoulder joints. 
These positions were indicated by color markers ($0.10$\;cm) placed on the arm (Fig.~\ref{setup}b), and were extracted using DLTdv8.  
An Inertial Measurement Unit (IMU) (Xsens DOT) affixed to the end-effector was used to capture acceleration data at $60$\;Hz.
To assess actuator performance we evaluated i) actuator pressure, ii) motion path length, iii) movement smoothness (straightness index (SI) and jerk) of the end effector, and iv) change in elbow joint angle~\cite{oguntosin2015development, zhang2008modeling} by different actuator variants. 
If an actuator did not demonstrate the highest level of performance, it was excluded from subsequent statistical analysis.

We hypothesized that actuator performance will be affected by number of cells, size, and shape.
Non-parametric statistical (Kruskal-Wallis) tests were performed to compare actuator performance based on different parameters (violation of normality was confirmed using the Kolmogorov-Smirnov test).
Post hoc Dunn’s pairwise comparisons with Bonferroni corrections were performed using SPSS v.27.

\subsection{Results}


\subsubsection{Pressure} 
\begin{figure}[!t]
\vspace{6pt}
     \centering
     \includegraphics[width=0.95\columnwidth]{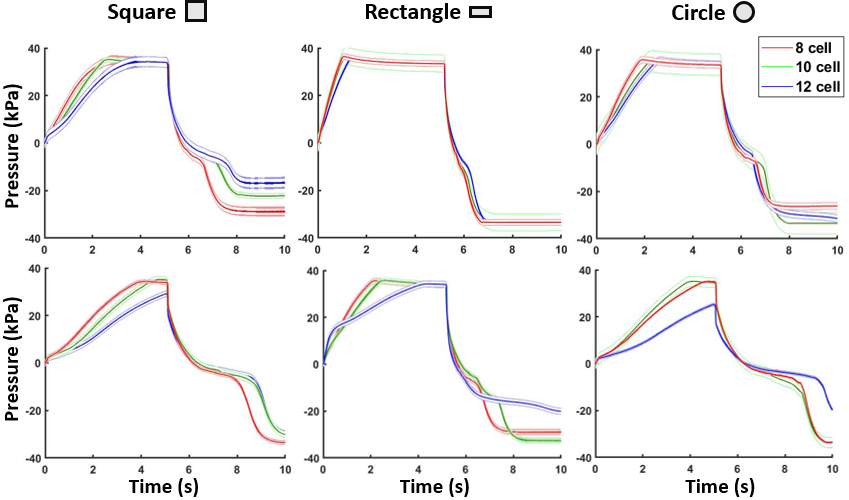}
     \vspace{-10pt}
         \caption{Variation in pressure profiles of $3$\;cm (Top) and $4$\;cm (Bottom) actuators of different cell shapes and numbers, observed during $5$\;sec inflation followed by $5$\;sec deflation on the infant model.}
     \label{pressure}
     \vspace{-20pt}
\end{figure}

When the actuator reaches full inflation/deflation, the pressure reaches a steady state. 
We observed an approximately $34-36$\;kPa pressure built within the actuators during full inflation. 
Based on the changes in the actuators' pressure over a period of time, each actuator's response time performance for flexion and extension of the elbow can be assessed. 
Irrespective of the shape of the actuator, the pressure profile suggests that it takes longer for the actuator to reach full inflation/deflation as the cell size and number increase (Fig.~\ref{pressure}). 
Four actuators (4cm-12-cell of all shapes and square 4cm-10-cell) were not able to fully inflate/deflate within $5$\;sec. 
Hence, these four actuators were unable to achieve their maximum performance within this specified time duration. 
These actuators might have a broader range of motion; however, considering a specific application with time constraints, these actuators might not be a suitable option. 
We also noticed that the rectangular shape is the fastest for inflation/deflation, due to its smaller footprint and cavity, compared with other shapes. 


\begin{figure}[!t]
\vspace{6pt}
\centering
   \includegraphics[width=0.95\linewidth]{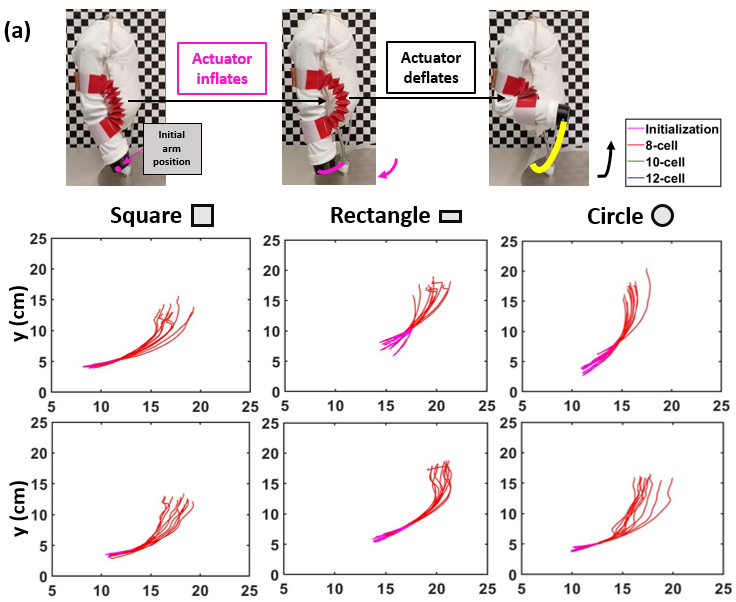}
   \includegraphics[width=0.95\linewidth]{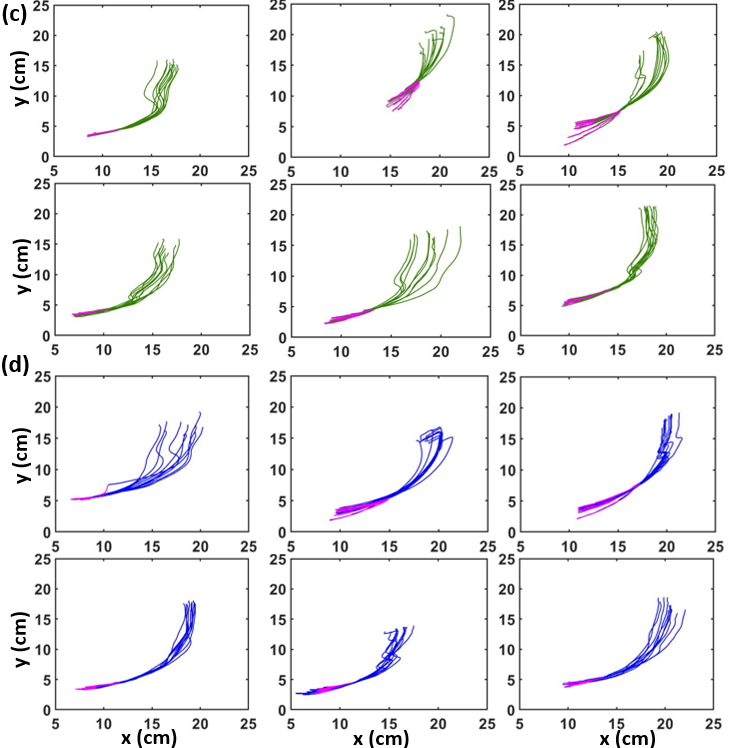}
\vspace{-10pt}
\caption{Motion path of the end-effector during inflation/deflation of various actuators. (a) The actuator maintains the arm at rest in an initial position. Inflation of the actuators brings the forearm to the minimum possible elbow bending angle, while deflation flexes the elbow to achieve the maximum possible bending angle. Trajectories for the 10 trials of (b), (c), and (d) show respectively 8-cell, 10-cell, and 12-cell actuators. The top row corresponds to 3cm actuators, and the bottom row corresponds to 4cm actuators.}
\label{path}
\vspace{-18pt}
\end{figure}

\subsubsection{Motion Path Length}
With the actuators inflated or deflated, the forearm moves as shown in Fig.~\ref{path}a. 
From Fig.~\ref{path}b, c, and d, the 2D position of the end-effector shows the performance variability of different actuators.
Actuators with greater cell numbers and larger dimensions elevate the forearm higher. 
However, due to incomplete inflation/deflation within $5$\;sec, the 4cm-12cell square and circle actuators did not achieve their optimal performance. 
Nevertheless, once they attain their peak state, they have the potential to further enhance elbow flexion.

Path lengths are listed in Table~\ref{path length}. 
A circular cell and a higher number of cells contributed to larger path lengths. 
No statistically significant difference in path length among the actuators was observed with varying cell size ($\chi\textsuperscript{2}(1)$=0.806, $p=0.369$) but a significant difference among the actuators with varying shape ($\chi\textsuperscript{2}(2)$=24.728, $p<0.001$) and cell numbers ($\chi\textsuperscript{2}(2)$=65.930, $p<0.001$) was observed. 
A posthoc analysis showed that the circular actuators (18.16$\pm$2.38 cm) have significantly higher ($p<0.001$) path length than square (15.09$\pm$1.91 cm)  and rectangular (16.04$\pm$3.33 cm) actuators, whereas the actuators with 8-cell (14.31$\pm$1.60 cm) have significantly lower ($p=0.000$)  path length than 10-cell (17.71$\pm$2.87 cm) and 12-cell (18.99$\pm$1.75 cm) actuators.

\begin{table}[!h]
    \vspace{6pt}
    \caption{Path length (cm)}
    \label{path length}
    \vspace{-12pt}
    \begin{center}
        \begin{adjustbox}{width=0.9\columnwidth,center}
            \begin{tabular}{c c c c c}
                \toprule
                \multirow{2}{*}{Size} & \multirow{2}{*}{Cell Number} & \multicolumn{3}{c}{Mean$\pm$ SD}  \\
                & &  Square & Rectangle & Circle \\
                \midrule
                \multirow{3}{*}{3 cm} & \multirow{1}{*}{8 cell} & 13.65$\pm$0.76 &12.22$\pm$0.74 &15.31$\pm$0.89\\
                & \multirow{1}{*}{10 cell} & 16.76$\pm$0.52 &12.82$\pm$1.15 &19.22$\pm$0.98\\
                & \multirow{1}{*}{12 cell} & 16.91$\pm$0.81 &20.06$\pm$1.20 &19.99$\pm$0.78\\
                \midrule
                \multirow{3}{*}{4 cm} & \multirow{1}{*}{8 cell} & 13.05$\pm$0.78 &15.97$\pm$0.67 &15.64$\pm$0.73\\
                & \multirow{1}{*}{10 cell} & 16.62$\pm$1.29* &19.13$\pm$0.46 &20.62$\pm$0.40\\
                & \multirow{1}{*}{12 cell} &20.53$\pm$0.57*  &17.76$\pm$0.72* &18.90$\pm$1.41*\\
                \bottomrule
            \end{tabular}
        \end{adjustbox}
    \end{center}
        *Not fully inflated/deflated
        \vspace{-20pt}
\end{table}






\subsubsection{Movement Smoothness}

Smoothness of motion for different actuators during elbow flexion was determined by the Straightness index (SI) and jerk (Table~\ref{smoothness}) computed by the end-effector 2D position and acceleration data from IMU. 
\paragraph{Straightness Index (SI)} SI is the ratio of actual path length to the shortest distance between the start and endpoint. 
Lower SI values imply that the trajectory is closer to being a straight line, while higher values indicate greater deviation and a less linear path. 
Here, the SI values of the motion path of the actuators are closer to 1, implying that the attained motions are smooth. 
An optimized motion might be possible with fewer cells, regardless of their size and shape, as no statistically significant difference in SI among the actuators was observed with varying shapes ($\chi\textsuperscript{2}(2)$=0.131, $p=0.937$) or size ($\chi\textsuperscript{2}(1)$=0.114, $p=0.735$) but revealed a significant difference among the actuators with varying cell numbers ($\chi\textsuperscript{2}(2)$=30.994, $p<0.001$).
A posthoc analysis showed that the 8-cell actuators (1.12$\pm$0.05) follow a significantly improved optimization path ($p=0.000$) than 10-cell (1.15$\pm$0.04) and 12-cell (1.22$\pm$0.12) actuators.

\paragraph{Jerk} 
High jerk values indicate abrupt changes in acceleration. 
Minimizing jerk can ensure smooth and natural movements, especially when interacting with infants to avoid sudden impacts or vibrations.
Jerk values here varied between 4.43$\pm$0.79 to 9.45$\pm$1.18 ms\textsuperscript{-3} for the functional actuators.
Smoothness of motion may not change based on the number of cells, but it was influenced by the size and shape of the cells. 
No statistically significant difference in jerk was observed among the actuators with varying cell numbers ($\chi\textsuperscript{2}(2)$=5.970, $p=0.051$) but revealed a significant difference among the actuators with varying shape ($\chi\textsuperscript{2}(2)$=10.849, $p=0.004$) and size ($\chi\textsuperscript{2}(1)$=25.337, $p<0.001$). 
Posthoc analysis showed that circular (5.87$\pm$2.02) and square (5.44$\pm$0.93) actuators are significantly smoother ($p=0.014$) than rectangle ones (6.67$\pm$1.22) while the $3$\;cm actuators (5.38$\pm$1.10) are significantly smoother ($p<0.001$) than the $4$\;cm ones (6.84$\pm$1.75).

\begin{table}[!h]
\vspace{6pt}
\caption{Smoothness Indices}
\label{smoothness}
\vspace{-12pt}
\begin{center}
\begin{adjustbox}{width=0.9\columnwidth,center}
\begin{tabular}{c c c c c}
\toprule
\multirow{2}{*}{Size} & \multirow{2}{*}{Cell Number} & \multicolumn{3}{c}{Straightness Index (SI) (Mean$\pm$ SD)}  \\
& &  Square & Rectangle & Circle \\
 \midrule
\multirow{3}{*}{3 cm} & \multirow{1}{*}{8 cell} & 1.12$\pm$0.03 &1.13$\pm$0.07 &1.09$\pm$0.03\\
 & \multirow{1}{*}{10 cell} & 1.16$\pm$0.02 &1.12$\pm$0.06 &1.17$\pm$0.05\\
 & \multirow{1}{*}{12 cell} & 1.14$\pm$0.02 &1.36$\pm$0.11 &1.16$\pm$0.03\\

 \midrule
\multirow{3}{*}{4 cm} & \multirow{1}{*}{8 cell} & 1.14$\pm$0.03 &1.15$\pm$0.08 &1.12$\pm$0.02\\
 & \multirow{1}{*}{10 cell} & 1.15$\pm$0.02* &1.14$\pm$0.02 &1.17$\pm$0.02\\
 & \multirow{1}{*}{12 cell} & 1.15$\pm$0.01* &1.40$\pm$0.09* &1.13$\pm$0.02*\\

\bottomrule
\vspace{1pt}
\end{tabular}
\end{adjustbox}

\begin{adjustbox}{width=0.9\columnwidth,center}
\begin{tabular}{c c c c c}
\toprule
\multirow{2}{*}{Size} & \multirow{2}{*}{Cell Number} & \multicolumn{3}{c}{Jerk (ms\textsuperscript{-3} (Mean$\pm$ SD)}  \\
& &  Square & Rectangle & Circle \\
 \midrule
\multirow{3}{*}{3 cm} & \multirow{1}{*}{8 cell} & 5.23$\pm$1.13 &6.16$\pm$0.96 &4.70$\pm$0.56\\
 & \multirow{1}{*}{10 cell} &5.08$\pm$0.62 &6.24$\pm$1.52 &4.78$\pm$0.25\\
 & \multirow{1}{*}{12 cell} &5.07$\pm$1.11 &6.20$\pm$0.55 &4.43$\pm$0.79\\

 \midrule
\multirow{3}{*}{4 cm} & \multirow{1}{*}{8 cell} & 5.87$\pm$0.53 &6.58$\pm$1.38 &9.45$\pm$1.18\\
 & \multirow{1}{*}{10 cell} & 4.22$\pm$0.67* &6.16$\pm$1.66 &6.03$\pm$0.53\\
 & \multirow{1}{*}{12 cell} & 5.57$\pm$0.79* &7.53$\pm$0.92* &7.17$\pm$1.09*\\

\bottomrule
\end{tabular}
\end{adjustbox}
\end{center}
        *Not fully inflated/deflated
        \vspace{-12pt}
\end{table}

\subsubsection{Elbow Joint Angle} 
\begin{figure}[!t]
\vspace{6pt}
     \centering
     \includegraphics[width=1\columnwidth]{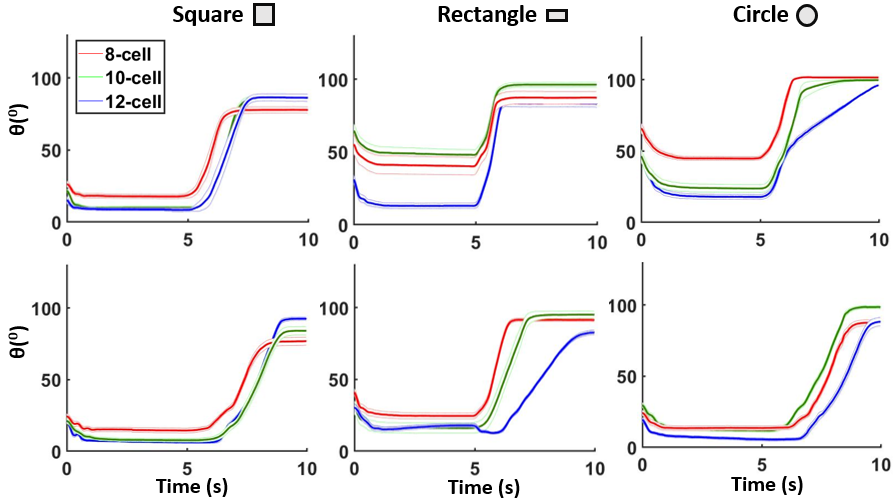}
     \vspace{-12pt}
         \caption{Change in elbow joint angle, $\theta$, over time. The top and bottom row corresponds to 3cm and 4cm actuators respectively.}
     \label{angle}
     \vspace{-18pt}
\end{figure}

    

When first placed on the top of the elbow joint, different actuator variants (unpowered) can keep the arm at rest at different elbow joint angles, $\theta$ (Fig.~\ref{angle}).
During inflation, the elbow extends and reaches a minimum $\theta\ge0$\;degrees, which will not force the arm beyond the natural range of motion. 
Next, during actuator deflation i.e. elbow flexion, the active ROM of the elbow is less than 105 degrees (except the circular ones).
The change in elbow joint angle during flexion was influenced by the shape, size, and number of the cells of the actuator.
A statistically significant difference in the actuators with varying cell numbers ($\chi\textsuperscript{2}(2)$=44.072, $p<0.001$), shape ($\chi\textsuperscript{2}(2)$=24.782, $p<0.001$), and size ($\chi\textsuperscript{2}(1)$=10.512, $p=0.001$) was observed. 
Posthoc analysis showed that circular actuators (75.14$\pm$10.04 degrees) perform significantly higher ($p=0.000$) than rectangular ones (63.25$\pm$12.99 degrees), while $4$\;cm actuators (74.78$\pm$7.80 degrees) perform significantly higher ($p=0.001$) than $3$\;cm ones (66.48$\pm$12.91 degrees).
8-cell actuators (62.67$\pm$9.46 degrees) showed significantly lower joint angles ($p=0.000$) than 10-cell (73.29$\pm$13.52 degrees) and 12-cell ones (76.60$\pm$5.14 degrees).

\section{Conclusion}
The paper addresses design and evaluation of a family of fabric-based bellow-type actuators for the infant elbow. 
Among 27 initially-proposed multi-cell designs, 18 were down-selected and tested on a physical infant-sized model.
Various considerations are key when varying the shape, size, and number of cells in this elbow actuator. 
For example, opting for a reduced number of cells can enhance the quality of the end-effector's motion path, while adopting smaller cell sizes can contribute to a smoother motion but may result in reduced flexion angles. 
Circular cells might facilitate less straight but longer motion paths and higher flexion angles at the elbow joint. 
These findings will inform the next steps in the design and control of our wearable device.


 

\bibliography{ipsita}
\bibliographystyle{IEEEtran}

\end{document}